\documentclass{article} 
\usepackage{iclr2020_conference,times}

\usepackage{amsmath,amsfonts,bm}

\def\eqref#1{equation~\ref{#1}}

\def\1{\bm{1}}

\DeclareMathAlphabet{\mathsfit}{\encodingdefault}{\sfdefault}{m}{sl}
\SetMathAlphabet{\mathsfit}{bold}{\encodingdefault}{\sfdefault}{bx}{n}

\usepackage{hyperref}
\usepackage{url}

\usepackage{subfigure}
\usepackage{multirow}
\usepackage{booktabs}
\usepackage{multicol}
\usepackage{verbatim}

\usepackage{algorithm}
\usepackage{algorithmicx}  
\usepackage{algpseudocode}
\usepackage{bm}
\usepackage{amssymb}
\usepackage[misc]{ifsym}
\usepackage{makecell}
\usepackage{appendix}
\usepackage{graphicx}

\newcommand\blfootnote[1]{%
  \begingroup
  \renewcommand\thefootnote{}\footnote{\hspace{-0.55cm}#1}%
  \addtocounter{footnote}{-1}%
  \endgroup
}

\title{Attacking and Defending Machine Learning Applications of Public Cloud\blfootnote{Accepted by Blackhat Asia 2020}}

\iclrfinalcopy

\author{Dou Goodman\textsuperscript{\Letter} , Hao Xin, Wang Yang \& Zhang Huan\\
Baidu X-Lab\\
Beijing, China\\
\texttt{\{liu.yan, haoxin01\}@baidu.com}
}

\begin{document}

\maketitle

\begin{abstract}
Adversarial attack breaks the boundaries of traditional security defense. For adversarial attack and the characteristics of cloud services, we propose Security Development Lifecycle for Machine Learning applications, e.g., SDL for ML. The SDL for ML helps developers build more secure software by reducing the number and severity of vulnerabilities in ML-as-a-service, while reducing development cost.
\end{abstract}

\section{Introduction}
In recent years, Machine Learning (ML) techniques have been extensively deployed for computer vision tasks, particularly visual classification problems, where new algorithms reported to achieve or even surpass the human performance\citep{Krizhevsky2012ImageNetCW,Girshick_2015,Najibi_2017}. Success of ML algorithms has led to an explosion in demand. To further broaden and simplify the use of ML algorithms, cloud-based services offered by Amazon, Google, Microsoft, Clarifai and other public cloud companies have developed ML-as-a-service tools. Thus, users and companies can readily benefit from ML applications without having to train or host their own models\citep{Hosseini_2017}. For example, Google introduced the Cloud Vision API for image analysis. A demonstration website has been also launched, where for any selected image, the API outputs the image labels, identifies and reads the texts contained in the image and detects the faces within the image. It also determines how likely is that the image contains inappropriate contents, including adult, spoof, medical, or violence contents. Unlike common attacks against web applications, such as SQL injection and XSS, there are very special attack methods for machine learning applications, e.g., \emph{Adversarial Attack}\citep{fischer2017adversarial,Xie_2017,wang2019daedalus,Carlini_2018,qin2019imperceptible,wang2019daedalus} and \emph{Spatial Attack}\citep{xiao2018spatially,goodman2019cloudbased,Li_2019}. Obviously, neither public cloud companies nor traditional security companies pay much attention to these new attacks and defenses\citep{goodman2019cloud,goodman2019cloudbased,goodman2020transferability}.

This paper focuses on the Cloud Vision API of public cloud companies and explores the attacks against the machine learning applications and describes effective defenses and mitigation. While the content is focused on the Cloud Vision API, some of the attack and defense topics are applicable to other machine learning applications such as Natural Language Processing (NLP) applications and speech processing applications. Our research involves attacks, intrusion detection, security testing and security reinforcement, which can become Security Development Lifecycle for Machine Learning applications, e.g., SDL for ML. 

Our key items covered:
\begin{itemize}
    \item FFL-PGD attack against image classification service
    \item Spatial attack against image search service
    \item Security testing for model robustness
    \item Securing machine learning applications against attacks
    \item Adversarial attack detection
\end{itemize}

\section{Adversarial Attack and Spatial Attack}
\subsection{Problem Formulation}
The function of a pre-trained classification model $F$, e.g. an image classification or image detection model, is mapping from input set to the label set. For a clean image example $O$, it is correctly classified by $F$ to ground truth label $y \in Y$, where $Y$ including $\{1,2, \ldots, k\}$ is a label set of $k$ classes. 
For the adversarial attack, an attacker aims at adding small perturbations in $O$ to generate adversarial example $ADV$, so that $F(ADV) \neq F(O)$, where $D(ADV,O)<\epsilon$, $D$ captures the semantic similarity between $ADV$ and $O$, $\epsilon$ is a threshold to limit the size of perturbations.
For the spatial attack, an attacker aims at making spatial transformation $T(\cdot)$ to generate adversarial example $T(O)$, so that $F(T(O)) \neq F(O)$.

\subsection{Threat Model}
We assume the attacker has black-box access to the target model: the attacker is not aware of the model architecture, parameters, or training data, and is only capable of querying the target model with supplied inputs and obtaining the output predictions and their confidence scores. We chose to use untargeted attack i.e., changing the model’s output, because it is more suitable as a benchmark method.

\subsection{Adversarial Attack}

Generating adversarial examples usually requires white-box access to the victim model, but the attacker can only access the APIs opened by cloud platforms\citep{goodman2019cloud}. Thus, keeping models in the cloud can usually give a (false) sense of security. Unfortunately, a lot of experiments have proved that attackers can successfully adversarial attack ML-as-a-service\citep{goodman2019cloudbased,goodman2020transferability}.

\textbf{Query-based Adversarial Attack} Query-based attacks are typical black-box attacks, attackers do not have the prior knowledge and get inner information of ML models through hundreds of thousands of queries to successfully generate an adversarial example \citep{Shokri2017Membership}. In \citep{ilyas2017query}, thousands of queries are required for low-resolution images. For high-resolution images, it still takes tens of thousands of times. For example, they achieves a 95.5\% success rate with a mean of 104,342 queries to the black-box classifier. In a real attack, the cost of launching so many requests is very high.

\textbf{Transfer Adversarial Attack} Transfer Adversarial Attack are first examined by \citep{szegedy2013intriguing}, which study the transferability between different models trained over the same dataset. \citep{Liu2016Delving} propose novel ensemble-based approaches to generate adversarial example and their approaches enable a large portion of targeted adversarial example to transfer among multiple models for the first time. It is a matter of luck to find an open source model with exactly the same functions as the target ML-as-a-service for a Transfer Adversarial Attack in a real attack.

\textbf{FFL-PGD Attack} Fast Feature map Loss PGD (FFL-PGD) Attack achieves a high bypass rate with a very limited number of queries. Instead of millions of queries in previous studies, FFL-PGD generates adversarial examples using only one or twe of queries\citep{goodman2020transferability}. 

The basic steps of FFL-PGD Attack are as follows:

\begin{enumerate}
	\item Shadow Model Training: the attacker queries the oracle with inputs selected by manual annotation to build a model $F{}'(x)$ approximating the oracle model $F(x)$ decision boundaries.
	\item Adversarial Sample Crafting: the attacker uses shadow model $F{}'(x)$  to craft adversarial samples, which are then misclassified by oracle $F(x)$ due to the transferability. 
\end{enumerate}	

Different from the previous work\citep{Papernot2016Practical}, FFL-PGD Attack proposes a special object function, which can reduce the difference of the low-level feature between the adversarial sample and the original image, and increase the difference of the high-level semantic feature. Experiments show that this strategy greatly improves the attack effect\citep{goodman2020transferability}.

\begin{figure*}[htbp]
	\subfigure[Escape Rate]{
		\label{fig:ffl:a}
		\begin{minipage}[t]{0.3\linewidth}
			\centering
			\includegraphics[width=1.5in]{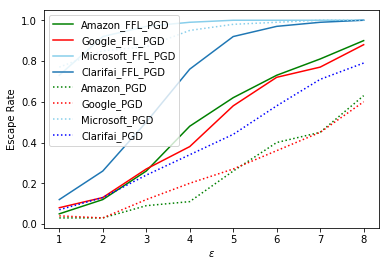}
		\end{minipage}
	}%
	\subfigure[PSNR]{
		\label{fig:ffl:b}
		\begin{minipage}[t]{0.3\linewidth}
			\centering
			\includegraphics[width=1.5in]{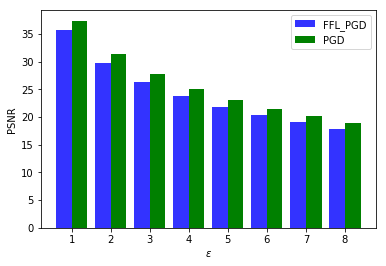}
		\end{minipage}
	}%
	\subfigure[SSIM]{
	\label{fig:ffl:c}
	\begin{minipage}[t]{0.3\linewidth}
		\centering
		\includegraphics[width=1.5in]{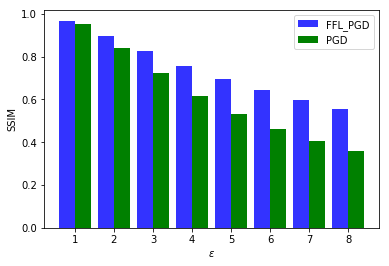}
	\end{minipage}
	}%
	\centering
	\caption{In (a), we increase step size $\epsilon$ from 1 to 8, the figure records the escape rates of PGD and FFL-PGD attacks against cloud-based image classification services under different $\epsilon$. In (b), the figure records the PSNR of PGD and FFL-PGD attacks and In (c), the figure records the SSIM of PGD and FFL-PGD attacks. }
	\label{fig:ffl}
\end{figure*}

The escape rates of PGD and FFL-PGD attacks are shown in Fig.~\ref{fig:ffl}. From Fig. \subref{fig:ffl:a}, we know that the ML-as-a-services of Amazon\footnote{https://aws.amazon.com/cn/rekognition/}, Google\footnote{https://cloud.google.com/vision/}, Microsoft\footnote{https://azure.microsoft.com} and Clarifai\footnote{https://clarifai.com} are vulnerable to PGD and FFL-PGD attacks. Step size $\epsilon$ controls the escape rate. Increasing this parameter can improve the escape rate. From Fig.~\subref{fig:ffl:b} and Fig.~\subref{fig:ffl:c}, we know that FFL-PGD attack has a success rate over 90\% among different ML-as-a-service and is considered acceptable for image quality and similarity.

\subsection{Spatial Attack}

\begin{table}[!htbp]
	\caption{Image processing methods commonly used in Spatial Attack.}
	\label{tab:Imageprocessing}
	\centering
	\resizebox{0.75\textwidth}{!}{
	\begin{tabular}{cl}
		\textbf{Image Processing Method} & \textbf{Literature} \\
		\midrule
		Gaussian Noise&\cite{Hosseini2017Google,li2019adversarial}\\ 
		Salt-and-Pepper Noise&\cite{Hosseini2017Google,li2019adversarial}\\ 
		Brightness Control&\cite{li2019adversarial,goodman2019cloudbased}\\ 
		Image Binarization&\cite{li2019adversarial}\\ 
		Grayscale Image&\cite{li2019adversarial,goodman2019cloudbased}\\ 
		Monochromatization&\cite{YuanStealthy,goodman2019cloudbased}\\
		Rotation&\cite{YuanStealthy,engstrom2017rotation}\\ 
		Texturing&\cite{YuanStealthy,goodman2019cloudbased}\\ 
		Blurring&\cite{YuanStealthy,goodman2019cloudbased}\\ 
		Transparentization \& overlap&\cite{YuanStealthy,engstrom2017rotation}\\ 
	\end{tabular}
	}
\end{table}

Spatial Attack can be understood as generalized Adversarial Attack. It does not affect human understanding of image content by transforming the original image, but it can fool the machine learning model. Different from Adversarial Attack, Spatial Attack usually affects all or most of the pixels, and human can perceive the changes of the image.

Prior work such as \citet{Hosseini2017Google} discussed Salt-and-Pepper Noise on Google vision APIs. \citet{YuanStealthy} report the first systematic study on the real-world adversarial images. 

As shown in the Table~\ref{tab:Imageprocessing}, we summarize several image processing methods commonly used in Spatial Attack. All these image processing techniques are implemented with Python libraries, such as skimage\footnote{https://github.com/scikit-image/skimage-tutorials} and OpenCV\footnote{https://opencv.org/}. Fig.~\ref{fig:spatial_attack_google_image_search} illustrates spoofing image search services with Spatial Attack.

\begin{figure*}[!hbpt]
	\centering
	\subfigure[Labeled as "cat"]{
		\label{fig:spatial_attack_google_image_search:a}
		\begin{minipage}{0.3\linewidth}
			\centering
			\includegraphics[width=2in]{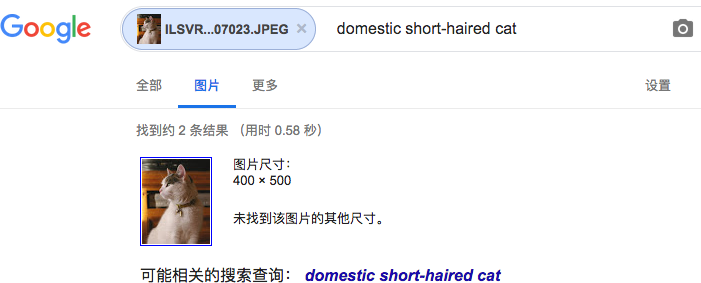}
		\end{minipage}
	}%
	\subfigure[Labeled as "flesh"]{
		\label{fig:spatial_attack_google_image_search:b}
		\begin{minipage}{0.3\linewidth}
			\centering
			\includegraphics[width=2in]{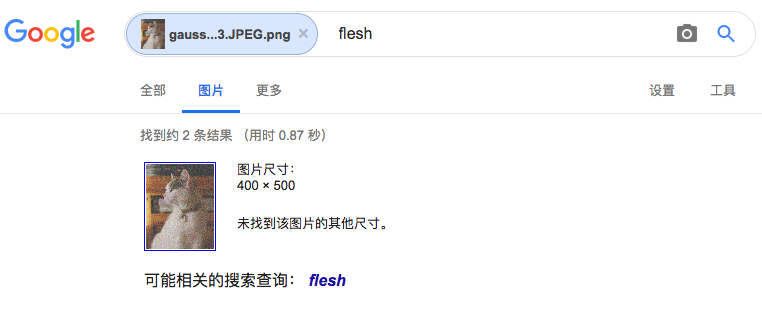}
		\end{minipage}
	}%
	\subfigure[Labeled as "rat"]{
		\label{fig:spatial_attack_google_image_search:c}
		\begin{minipage}{0.3\linewidth}
			\centering
			\includegraphics[width=2in]{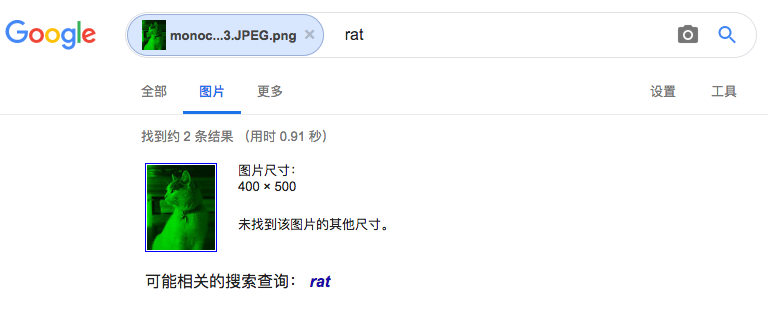}
		\end{minipage}
	}%
	
	\centering
	\caption{Illustration of the Spatial Attack against Google Images Search. (a) is origin image, search result is a cat and (b) is Gaussian Noise, search result is a "flesh", (c) is Monochromatization, search result is a "rat". }
	\label{fig:spatial_attack_google_image_search}
\end{figure*}

\section{SDL for ML}
\subsection{Overview}

Microsoft\footnote{https://www.microsoft.com/} has proposed SDL for traditional software development and provides developers with a lot of best practices\footnote{https://www.microsoft.com/en-us/securityengineering/sdl/practices}. Adversarial attack breaks the boundaries of traditional security defense. For adversarial attack and the characteristics of cloud services, We propose Security Development Lifecycle for Machine Learning applications, e.g., SDL for ML.

The SDL for ML helps developers build more secure software by reducing the number and severity of vulnerabilities in ML-as-a-service, while reducing development cost. 

\begin{table*}[!htbp]
	\caption{Overview of SDL for ML.}
	\label{tab:Overview}
	\centering
	\resizebox{0.9\textwidth}{!}{
	\begin{tabular}{cl}
		\textbf{Stages of Software Development} & \textbf{Components of SDL for ML }\\
		\midrule
		Design &  \makecell*[l]{Provide Training (details in Section~\ref{sec:ProvideTraining}) \\ \&  Establish Design Requirements (details in Section~\ref{sec:EstablishDesignRequirements}) }\\ 
		Coding & Adversarial Attack Mitigation (details in Section~\ref{sec:AdversarialAttackMitigation})\\ 
		Test & Robustness Evaluation Test (details in Section~\ref{sec:RobustnessEvaluationTest})\\ 
		Product release & N/A\\
		Operation and maintenance & Adversarial Attack Detection (details in Section~\ref{sec:AdversarialAttackDetection})\\ 
	\end{tabular}
	}
\end{table*}

\subsection{Provide Training}
\label{sec:ProvideTraining}

The safety awareness of employees and the training of using safety tools is a very important part of SDL. A software development enterprise usually has the following roles: RD is the software developer, QA is the software tester, OP is responsible for the operation and maintenance, and the training they need covers at least the following aspects detailed as Table~\ref{tab:Training}.

\begin{table*}[!htbp]
	\caption{Overview of Provide Training.}
	\label{tab:Training}
	\centering
	\resizebox{0.9\textwidth}{!}{
	\begin{tabular}{cl}
		\textbf{Training Contents} & \textbf{Roles}\\
		\midrule
		What is Adversarial Attack? What are the corresponding hazards?& RD \& QA \& OP\\
		How to conduct Adversarial Training? How to blur?& RD\\ 
		How to evaluate test robustness ?&  QA \& RD\\ 
		How to detect Adversarial Attack? & OP \& our \\
	\end{tabular}
	}
\end{table*}

\subsection{Establish Design Requirements}
\label{sec:EstablishDesignRequirements}

The SDL for ML is typically thought of as assurance activities that help engineers implement “secure features” of Adversarial Attack, in that the features are well engineered with respect to security. To achieve this, engineers will typically rely on security features, such as Blurring, Adversarial Training, and others. 

\subsection{Adversarial Attack Mitigation}
\label{sec:AdversarialAttackMitigation}

\begin{figure}[htbp] 
	\centering 
	\includegraphics[width=0.6\textwidth]{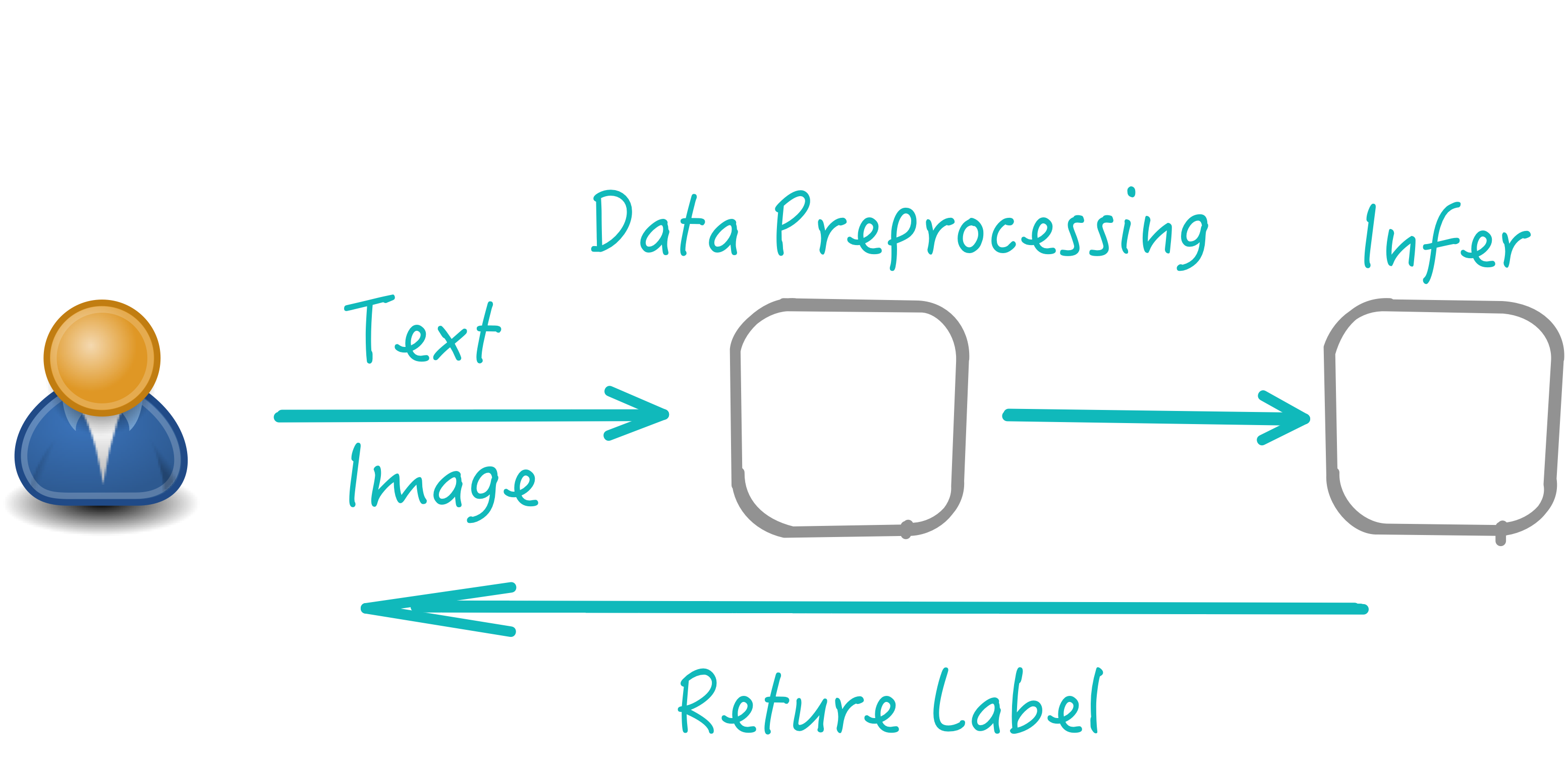}
	\caption{Dataflow of ML-as-a-service. } 
		\label{fig:cloud}
\end{figure}

Adversarial Attack Mitigation have two types of defense strategies\citep{yuan2017adversarial}: 
\begin{itemize}
    \item Reactive: detect adversarial examples after deep neural networks are built, e.g., Adversarial Detecting\citep{lu2017safetynet}, Input Reconstruction\citep{meng2017magnet}, and Network Verification\citep{katz2017reluplex}.
    \item Proactive: make deep neural networks more robust before adversaries generate adversarial examples, e.g., Network Distillation\citep{papernot2016distillation}, Adversarial training\citep{Madry2017Towards}, and Classifier Robustifying\citep{bradshaw2017adversarial}.
\end{itemize}

\citet{obfuscated-gradients} evaluate the robustness of nine papers \citep{buckman2018thermometer,ma2018characterizing,guo2017countering,dhillon2018stochastic,Xie2017Mitigating,song2017pixeldefend,samangouei2018defense,Madry2017Towards,na2017cascade} accepted to ICLR 2018 as non-certified white-box-secure defenses to adversarial examples,  they find that seven of the nine defenses use obfuscated gradients, a kind of gradient masking or input reconstruction, as a phenomenon that leads to a false sense of security in defenses against adversarial examples. Obfuscated gradients provide a limited increase in robustness and can be broken by improved attack techniques they develop. \citet{obfuscated-gradients} show that \textbf{the only defense significantly increases robustness to adversarial examples within the threat model proposed is adversarial training}. 


Considering the realization difficulty and actual effect, we recommend using Input Reconstruction in the data preprocessing stage and adversarially trained models in the model prediction stage. Adversarial Attack Mitigation Methods for ML-as-a-service are detailed in Table~\ref{tab:Recommended}, and we recommend Blurring\citep{Hosseini2017Google} and PGD Adversarial Training\citep{Madry2017Towards}.


\begin{table*}[!htbp]
	\caption{Adversarial Attack Mitigation Methods for ML-as-a-service.}
	\label{tab:Recommended}
	\centering
	\resizebox{0.9\textwidth}{!}{
	\begin{tabular}{cl}
		\textbf{Stages of ML-as-a-service } & \textbf{Mitigation Methods }\\
		\midrule
		Input Preprocessing&\makecell*[l]{Feature Squeezing \& Spatial Smoothing\citep{xu2017feature}\\
		Randomization\citep{Xie2017Mitigating}\\
		Blurring\citep{Hosseini2017Google}\\
		}\\ 
		\hline
		Prediction&\makecell*[l]{PGD Adversarial Training\citep{Madry2017Towards}\\
		                         Gaussian Augmentation\citep{zantedeschi2017efficient}\\
		                         Ensembling Adversarial Training\citep{tramer2017ensemble}\\
		                         Adversarial Logit Pairing\citep{DBLP:journals/corr/abs-1803-06373}\\
		                         Regularizing Input Gradients\citep{Ross2017Improving}\\
		                         Randomized Adversarial Training\citep{Araujo2019Robust}\\
		                         Feature Denoising\citep{Xie2018Feature}\\
		                         Attention and Adversarial Logit Pairing\citep{goodman2019improving}
		                         }\\ 
	\end{tabular}
	}
\end{table*}

\citet{hendrycks2019using,zheng2019efficient,davchev2019empirical} show that pre-training can improve model robustness and uncertainty. Therefore, using adversarially trained models on the ImageNet dataset for transfer learning should be a best practice.

Adversarial training included adversarial examples in the training stage and generated adversarial examples in every step of training and inject them into the training set. On the other hand, we can also generate adversarial samples offline, the size of adversarial samples is equal to the original data set, and then retrain the model. We have developed AdvBox\citep{goodman2020advbox}\footnote{https://github.com/advboxes/AdvBox}, which is convenient for developers to generate adversarial samples quickly. 

\begin{table*}[!htbp]
	\caption{Baseline attack methods of AdvBox.}
	
	\label{tab:Overview}
	\centering
	\resizebox{0.75\textwidth}{!}{
	\begin{tabular}{cl}
		\textbf{$L_p$ Attack} & \textbf{Baseline Attack Methods} \\
		\hline
		$L_0$ Attack & JSMA\citep{papernot2016limitations}\\ 
		$L_2$ Attack & CW\citep{carlini2017towards}\\ 
		$L_{\infty}$ Attack & FGSM\citep{goodfellow2014explaining} \& PGD\citep{aleks2017deep}\\ 
	\end{tabular}
	}
\end{table*}

\begin{table*}[htbp]
	\caption{Methods and parameters of defenses during the training and image preprocessing phase.}
	\label{tab:defense_parameters}
	\centering
	\begin{tabular}{cll} 
		\textbf{Stage} & \textbf{Method} & \textbf{Parameters} \\
		\midrule
		\multirow{6}{*}{Training}& Random Rotation(degree range)&  (0,360)\\
						  & Random Grayscale(probability) & 0.5\\
						  & Random Horizontal Flip(probability) & 0.5\\
						  & Random Resize and Crop(image size) & 224\\
						  & Gauss Filter(ksize) & 29\\
						  & Median Filter(ksize) & 11\\
		\midrule
		\multirow{2}{*}{Image preprocessing}&Median Filter(ksize) & 11\\
						 & Grayscale & N/A\\
	\end{tabular}
\end{table*}

\begin{table}[hbtp]
	\caption{Defense rates of Spatial Attack. Our Adversarial Training can raise the defense rate to more than 80\%, we have used the black line to thicken it.}
	\label{tab:discussion_defense}
	\centering
	\setlength{\tabcolsep}{1mm}{
	\begin{tabular}{cll}
		\textbf{Attack} & \textbf{w/o Defense}& \textbf{w/ Defense} \\
		\midrule
		Gaussian Noise & 0.60&\textbf{0.80}\\
		Rotation &0.70&\textbf{0.80}\\
		Salt-and-Pepper Noise &0.50&\textbf{0.95} \\
		Monochromatization & 0.4&\textbf{0.80} \\
	\end{tabular}
}
\end{table}

Table~\ref{tab:defense_parameters} and Table~\ref{tab:discussion_defense} show that our defense technology can effectively resist known Spatial Attack, such as Gaussian Noise, Salt-and-Pepper Noise, Rotation, and Monochromatization. 

\subsection{Robustness Evaluation Test}
Using multiple methods, according to certain conditions (method thresholds) to generate examples with a limited visual difference from the original image, but it is possible to make the model predict the wrong labels to evaluate the robustness of the model in these environments. It can be of two types: the first is safety-related, using adversarial examples formed by spatial transformation or image corruption, such as scaling, light transformation, weather, blur, shake, etc. The second is security-related, which uses the model gradient to stack perturbation to attack, such as FGSM,PGD,C/W, etc. The first is more general and more common, and also supports black-box testing. The second is more targeted. The human eye is less likely to detect it, but it relies more on white-box attacks. It is very difficult for an attacker to obtain model parameters, so the safety-related robustness test is more practical.

Currently, we provide open-source versions of robust tools for such evaluations: perceptron-benchmark\footnote{https://github.com/advboxes/perceptron-benchmark/}. The tool supports the testing of local models and cloud APIs. It uses 15 evaluation methods such as brightness, contrast, rotation, noise, shake, occlusion, frost, rain, fog, and snow, etc. Because each method can set different thresholds, and the degree of image corruption and attack effect are also different, we use PSNR and SSIM as auxiliary evaluation standards. The images generated by each method must ensure that the PSNR and SSIM are within a reasonable range, to ensure that the formed corruption is within the acceptable range of the human eye.

The original image for robustness testing can be generated using the test sets of the corresponding model. The results need to be evaluated after the robustness tests of these methods are completed. Analyze the weak points of the model and reasonable improvement methods. The following uses the Pytorch model InceptionV3 of the ImageNet1000 dataset as an example. We use ImageNet's validation sets as our test dataset, a total of 50,000 images. and 13 methods to test robustness. You can see that the accuracy of the model is as follows:

\begin{table*}[!htbp]
    \caption{Model accuracy under several perturbations}
    \resizebox{0.9\textwidth}{7mm}{
    \begin{tabular}{c|c|ccccccc}
    \toprule  
Network&original&gaussian\_noise&brightness&contrast &gaussian\_blur&rotation&raining &snowing\\
\midrule  
InceptionV3& \textbf{77\% }& 52\% &   60\% & 55\%   &   20\% &  30\%     &  51\%  &  40\%\\
    \bottomrule 
    \end{tabular}}
	
\end{table*}

\begin{figure*}[!hbpt]
	\centering
	\subfigure[$noise$]{
		\label{fig:bird:a}
		\begin{minipage}[t]{0.25\linewidth}
			\centering
			\includegraphics[width=1.2in]{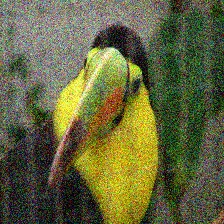}	
		\end{minipage}%
	}%
	\subfigure[$brightness$]{
		\label{fig:bird:b}
		\begin{minipage}[t]{0.25\linewidth}
			\centering
			\includegraphics[width=1.2in]{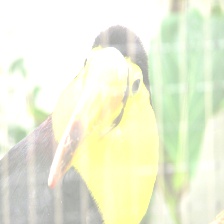}	
		\end{minipage}%
	}%
	\subfigure[$contrast$]{
		\label{fig:bird:c}
		\begin{minipage}[t]{0.25\linewidth}
			\centering
			\includegraphics[width=1.2in]{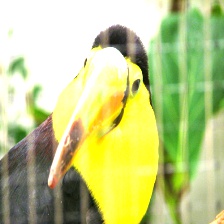}
			
		\end{minipage}
	}%
	\subfigure[$blur$]{
		\label{fig:bird:d}
		\begin{minipage}[t]{0.25\linewidth}
			\centering
			\includegraphics[width=1.2in]{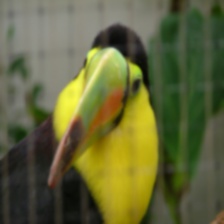}
			
		\end{minipage}
	}%
	
	\subfigure[$rotation$]{
		\label{fig:bird:e}
		\begin{minipage}[t]{0.25\linewidth}
			\centering
			\includegraphics[width=1.2in]{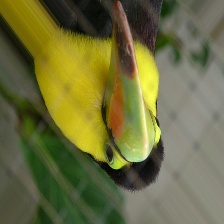}
			
		\end{minipage}
	}%
	\subfigure[$raining$]{
		\label{fig:bird:f}
		\begin{minipage}[t]{0.25\linewidth}
			\centering
			\includegraphics[width=1.2in]{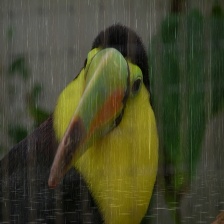}
			
		\end{minipage}
	}%
	\subfigure[$snowing$]{
		\label{fig:bird:g}
		\begin{minipage}[t]{0.25\linewidth}
			\centering
			\includegraphics[width=1.2in]{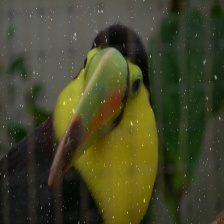}
			
		\end{minipage}
	}%

	\centering
	\caption{Effect of each type of corruption}
	\label{fig:spatial_attack_google_image_search}
\end{figure*}
 The Top-1 accuracy of the original image is 77.294\%. The results of the robustness test show that after the image is moderately corrupted, the model's classification accuracy rate has decreased significantly, with a maximum decrease of 50\% +, but the human eye can still correctly judge, indicating that the safety-related robustness from the black-box is still very fragile. and for filtered (blurred) pictures, the model's anti-interference ability is the weakest.

Specific instructions:
\begin{itemize}
    \item In the angle rotation performance, especially the positive and negative 135 degrees have an attack success rate of close to 70\%, which indicates that the confrontation at this angle is the weakest, and this type of processing can be added to the training sets to make up.
    \item In the performance of noise, most of the labels after the attack focus on several categories, which is particularly obvious in the salt and pepper noise. It shows that these categories are easy to attack in the model, and these labels of training data can be added accordingly.
    \item The blur corruption type has a high PSNR value, which is basically above 20, and the success rate of the attack is very high. For example, the success rate of the Gaussian filter attack is 80\%, and the average filter is 65\%. It shows that using the blur method to attack this model has both good clarity and a high success rate.
\end{itemize}

These weaknesses of the model that have been demonstrated through robustness testing can be targeted to use the adversarial training as mentioned before to compensate for it.

\label{sec:RobustnessEvaluationTest}

\subsection{Adversarial Attack Detection}
\label{sec:AdversarialAttackDetection}

\begin{figure}[htbp] 
	\centering 
	\includegraphics[width=0.8\textwidth]{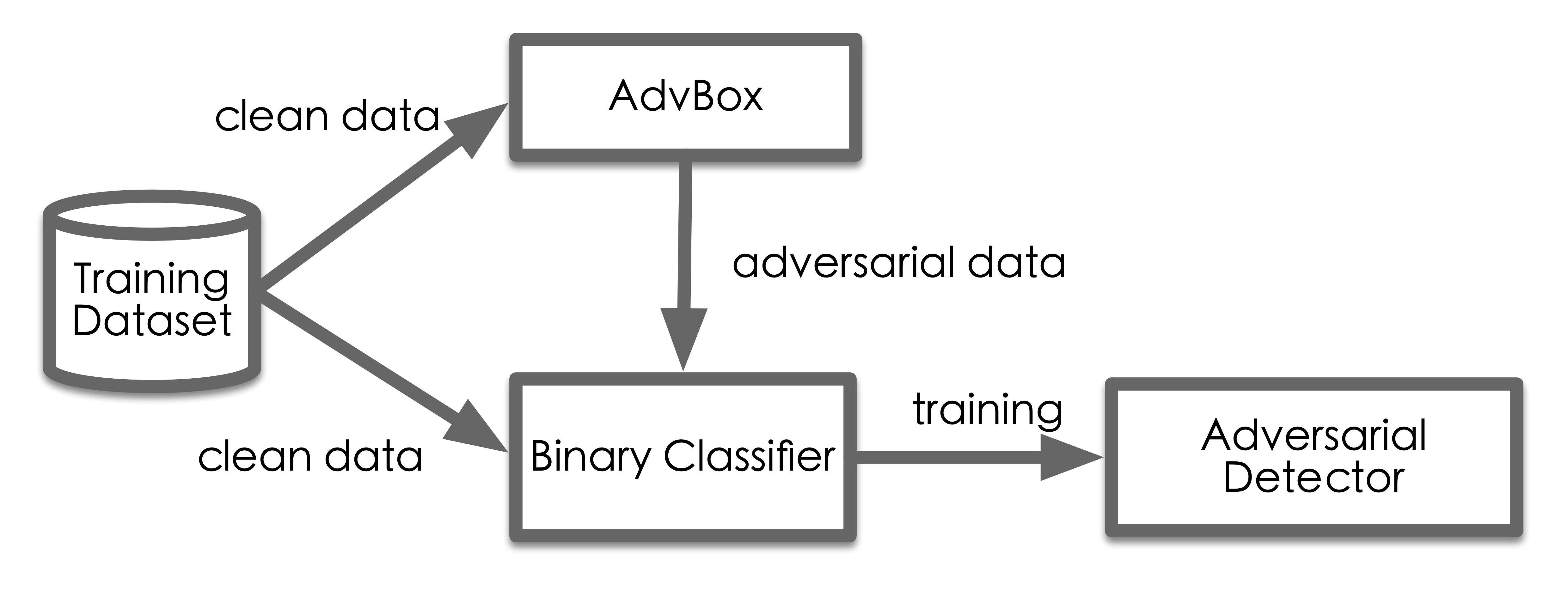}
	\caption{Dataflow of training deep neural network-based binary classifiers as detectors to classify the input data as a legitimate (clean) input or an adversarial example.} 
		\label{fig:detector}
\end{figure}

Adversarial example is essentially a kind of data, so a natural idea is: training deep neural network-based binary classifiers as detectors to classify the input data as a legitimate (clean) input or an adversarial example\citep{metzen2017detecting,bhagoji2017dimensionality,feinman2017detecting,grosse2017statistical}. See Fig.~\ref{fig:detector} a for details.


\section{Conclusion}
For adversarial attack and the characteristics of cloud services, we propose Security Development Lifecycle for Machine Learning applications, e.g., SDL for ML. The SDL for ML helps developers build more secure software by reducing the number and severity of vulnerabilities in ML-as-a-service, while reducing development cost. Provide Training, Establish Design Requirements, Adversarial Attack Mitigation, Robustness Evaluation Test, Adversarial Attack Detection are included in SDL for ML, and most of the features are already supported in our AdvBox.

\bibliography{iclr2020_conference,public}
\bibliographystyle{iclr2020_conference}

\newpage
\appendix
\section*{Appendix}
\section{Illustration of Spatial Attack }
\begin{figure*}[!h]
	\centering
	\subfigure[Origin]{
		\label{fig:cat:a}
		\begin{minipage}[t]{0.25\linewidth}
			\centering
			\includegraphics[width=1.2in]{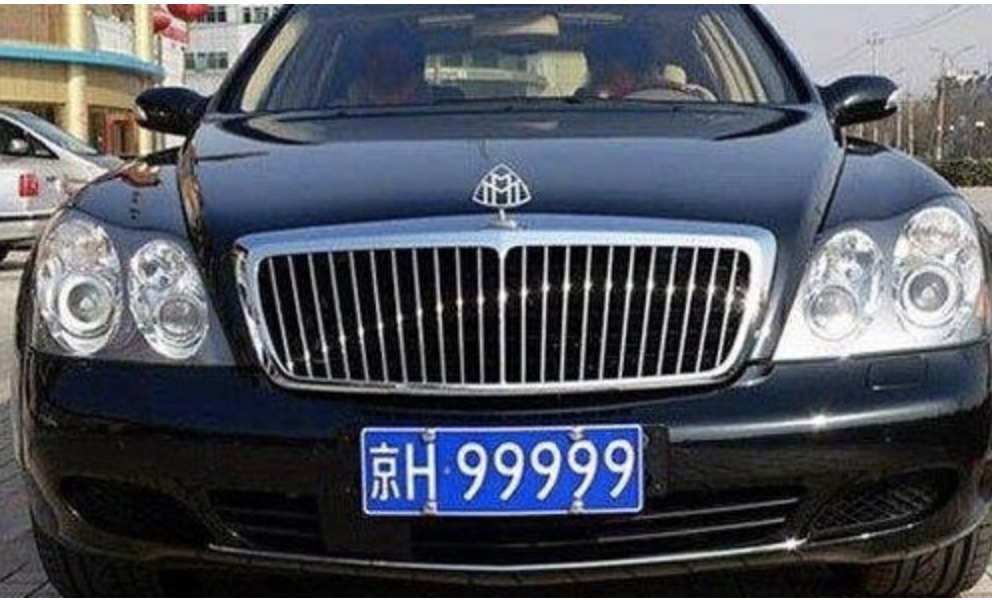}	
		\end{minipage}%
	}%
	\subfigure[$Gaussian$]{
		\label{fig:cat:b}
		\begin{minipage}[t]{0.25\linewidth}
			\centering
			\includegraphics[width=1.2in]{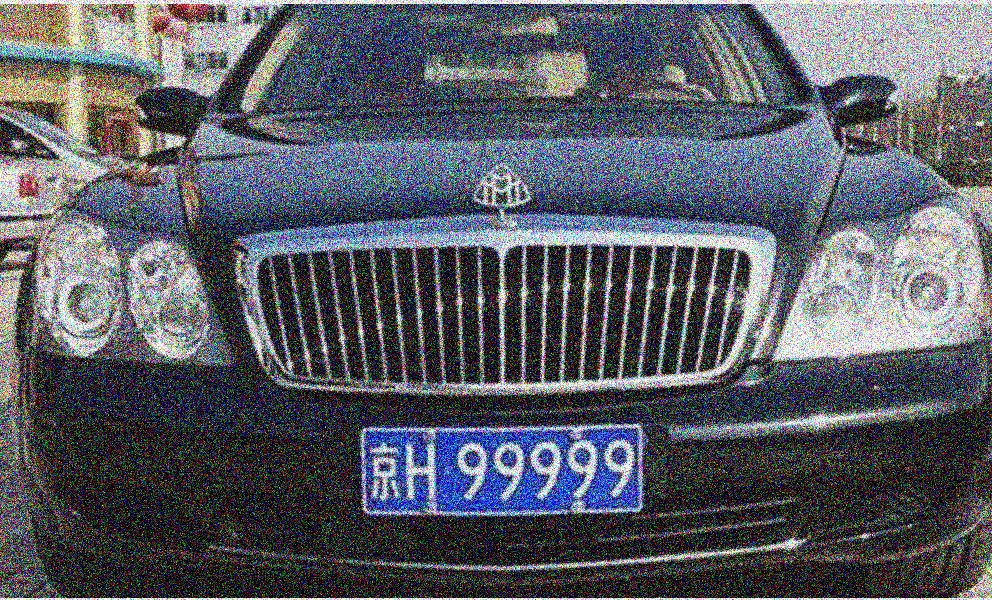}	
		\end{minipage}%
	}%
	\subfigure[$S\&P$]{
		\label{fig:cat:c}
		\begin{minipage}[t]{0.25\linewidth}
			\centering
			\includegraphics[width=1.2in]{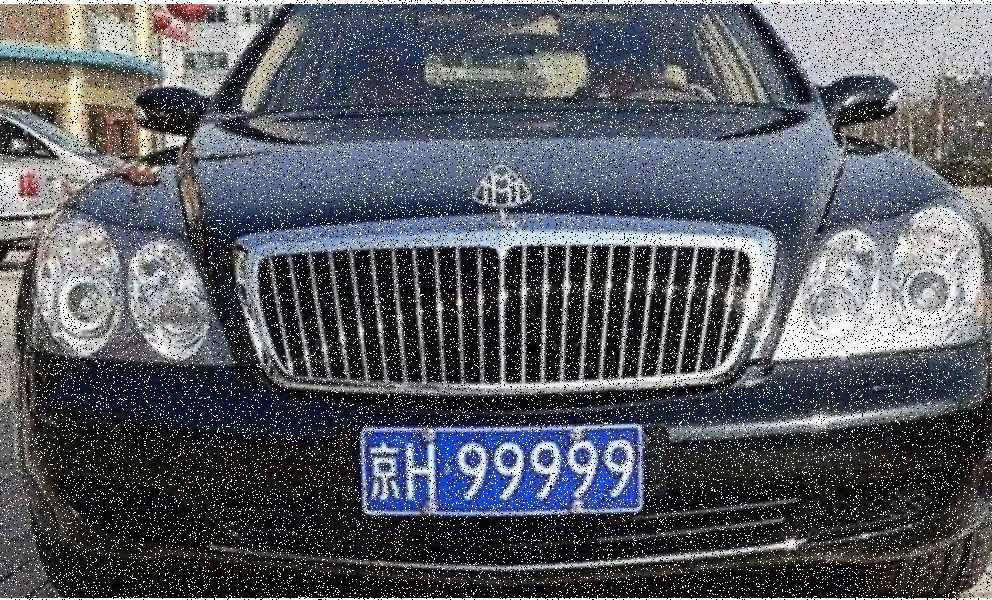}
			
		\end{minipage}
	}%
	\subfigure[$Rotation$]{
		\label{fig:cat:d}
		\begin{minipage}[t]{0.25\linewidth}
			\centering
			\includegraphics[width=1.2in]{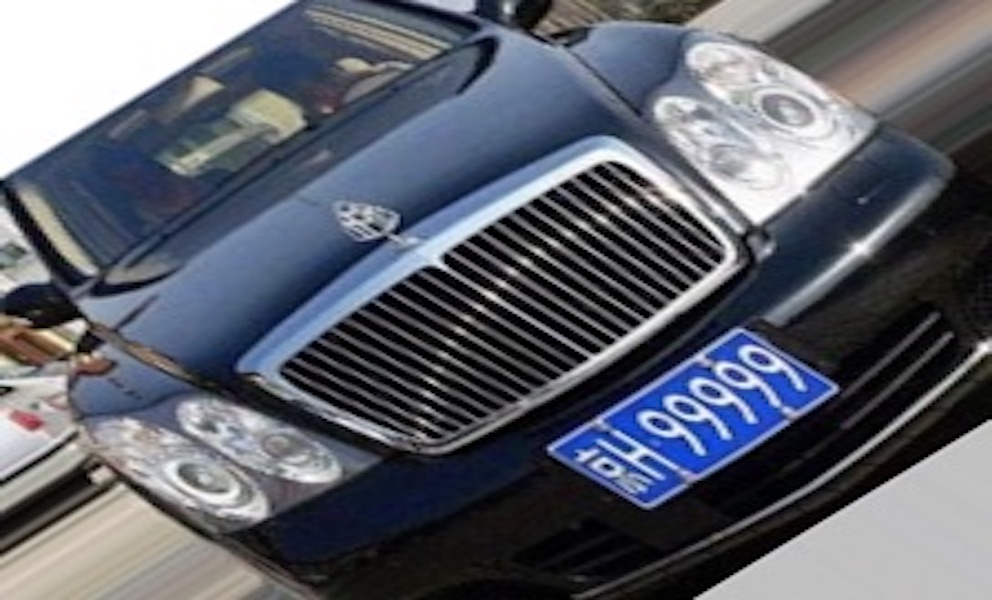}
			
		\end{minipage}
	}%
	
	\subfigure[$color=gray$]{
		\label{fig:cat:e}
		\begin{minipage}[t]{0.25\linewidth}
			\centering
			\includegraphics[width=1.2in]{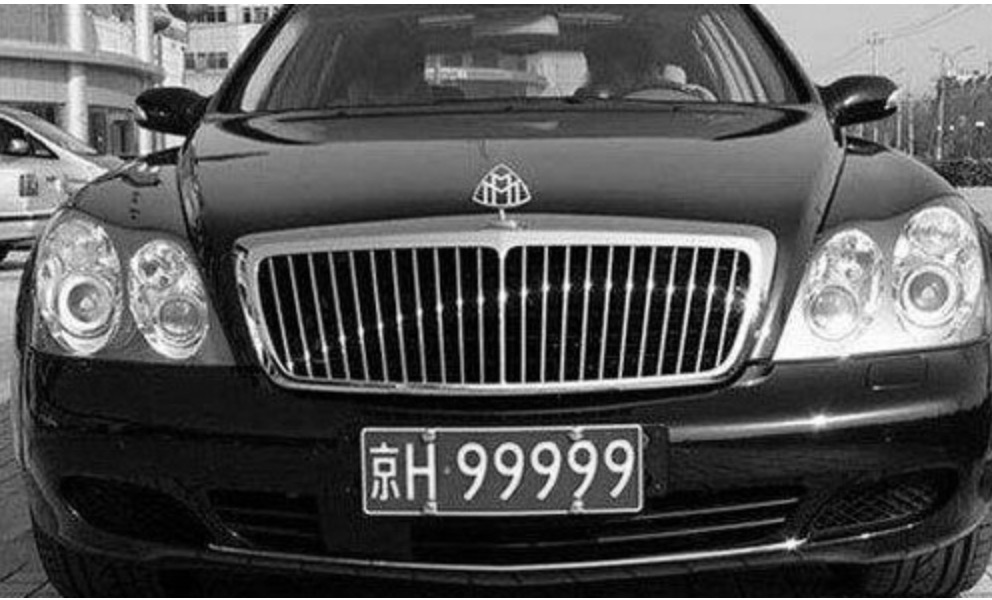}
			
		\end{minipage}
	}%
	\subfigure[$color=green$]{
		\label{fig:cat:f}
		\begin{minipage}[t]{0.25\linewidth}
			\centering
			\includegraphics[width=1.2in]{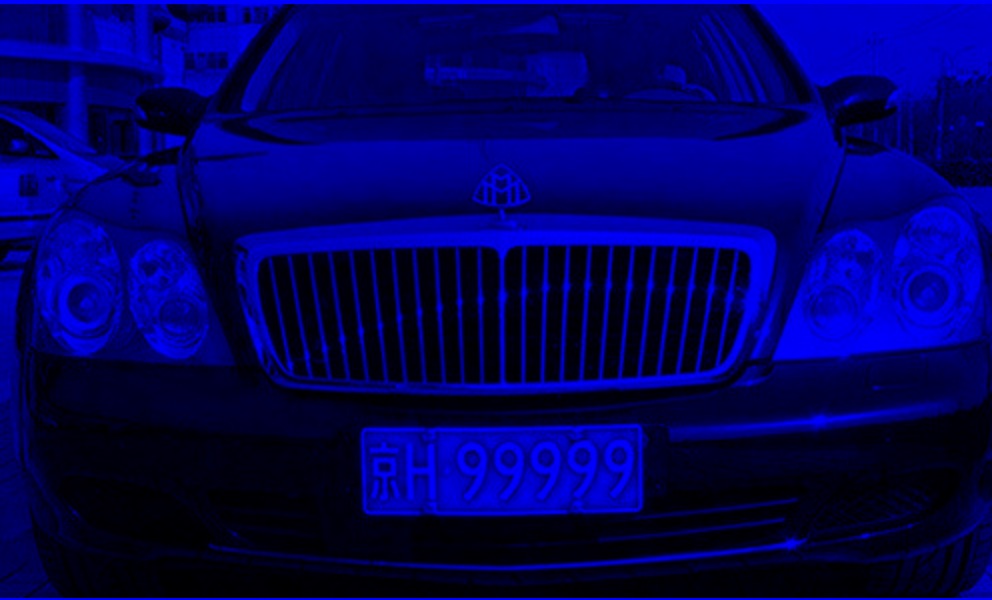}
			
		\end{minipage}
	}%
	\subfigure[$color=red$]{
		\label{fig:cat:g}
		\begin{minipage}[t]{0.25\linewidth}
			\centering
			\includegraphics[width=1.2in]{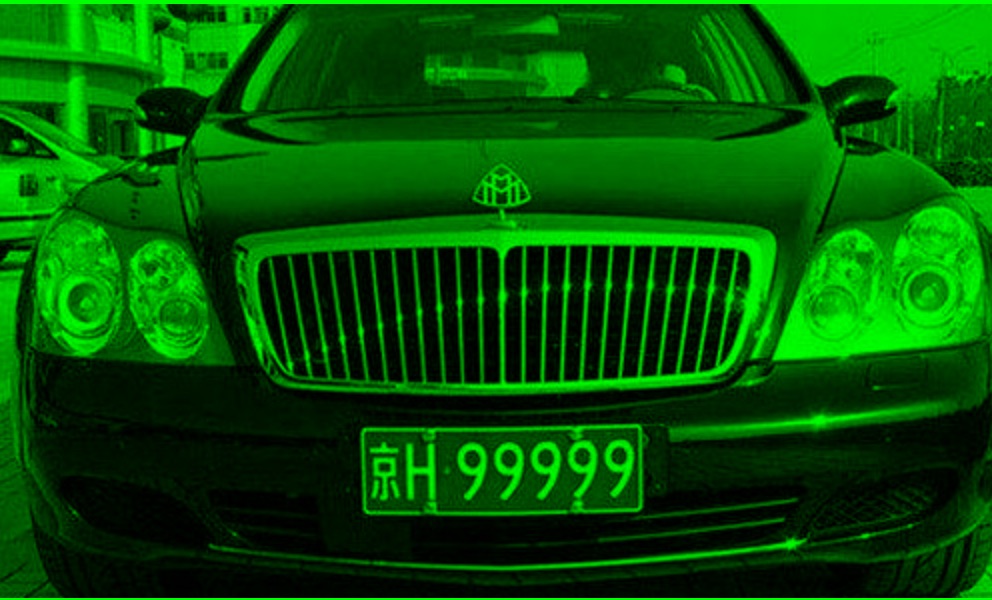}
			
		\end{minipage}
	}%
	\subfigure[$color=blue$]{
		\label{fig:cat:h}
		\begin{minipage}[t]{0.25\linewidth}
			\centering
			\includegraphics[width=1.2in]{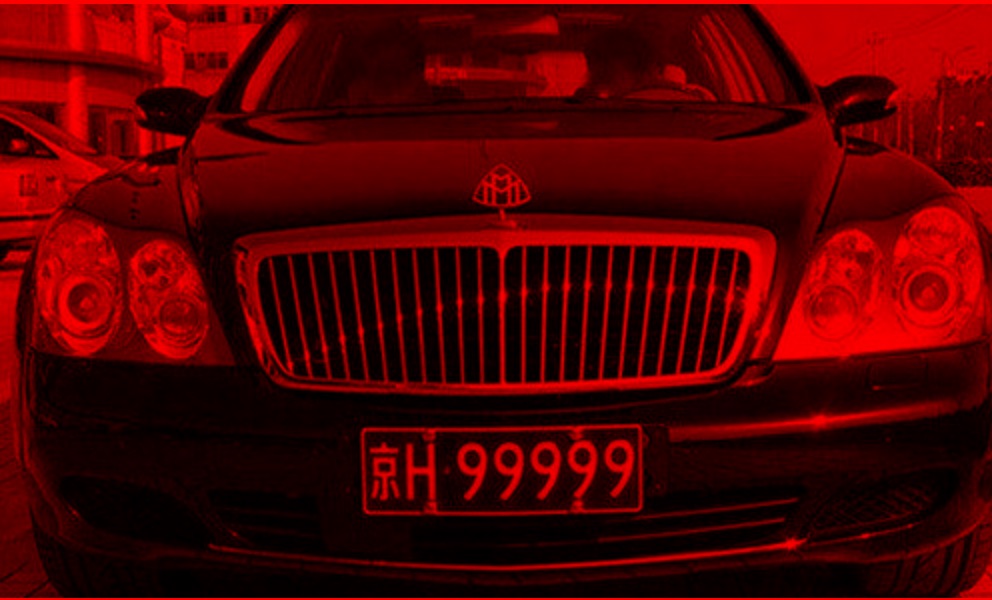}
			
		\end{minipage}
	}%
	
	\subfigure[$Blur$]{
		\label{fig:cat:i}
		\begin{minipage}[t]{0.25\linewidth}
			\centering
			\includegraphics[width=1.2in]{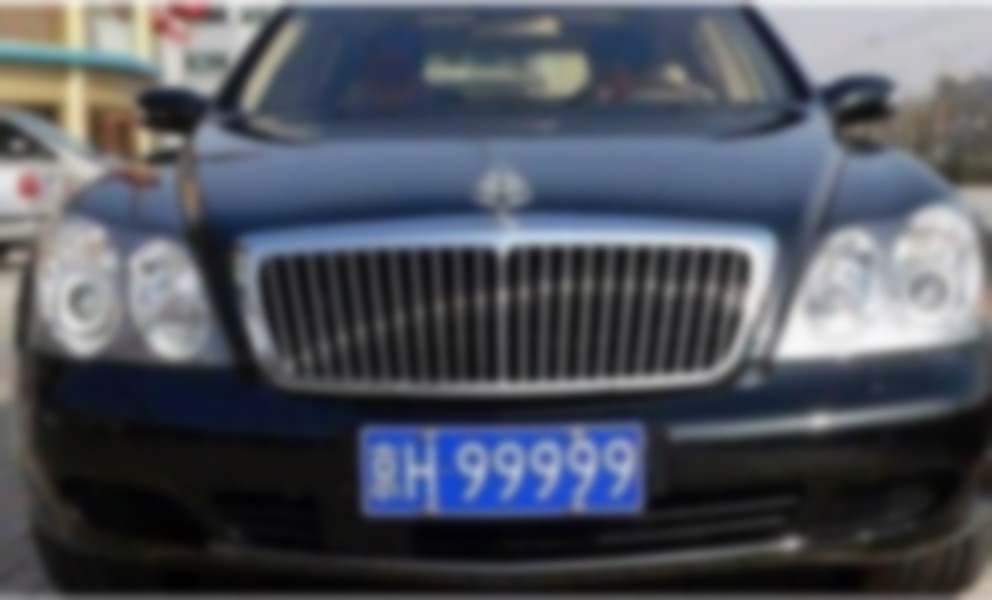}
			
		\end{minipage}
	}%
	\subfigure[$Frost$]{
		\label{fig:cat:j}
		\begin{minipage}[t]{0.25\linewidth}
			\centering
			\includegraphics[width=1.2in]{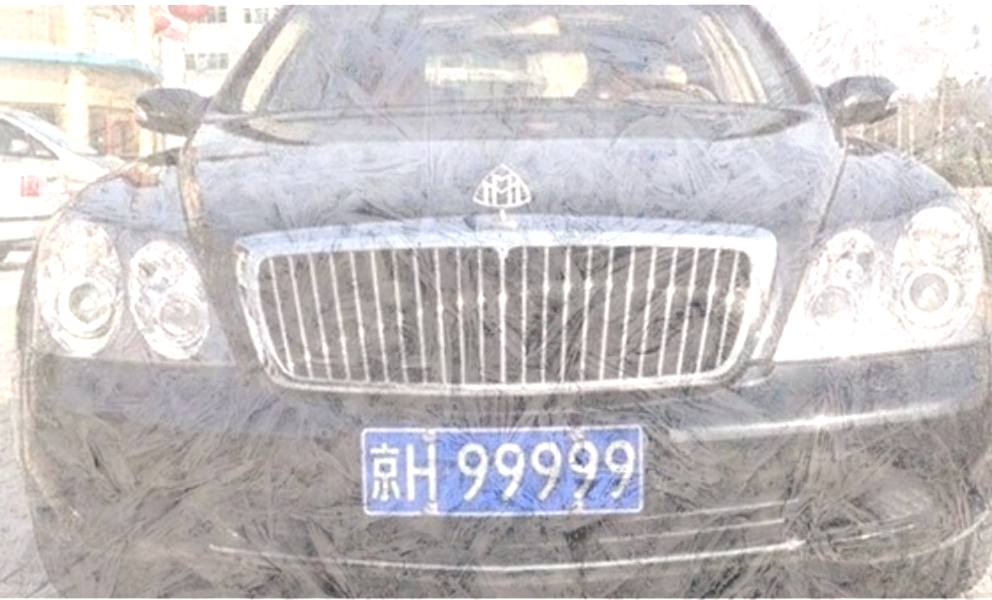}
			
		\end{minipage}
	}%
	\subfigure[$Flares$]{
		\label{fig:cat:k}
		\begin{minipage}[t]{0.25\linewidth}
			\centering
			\includegraphics[width=1.2in]{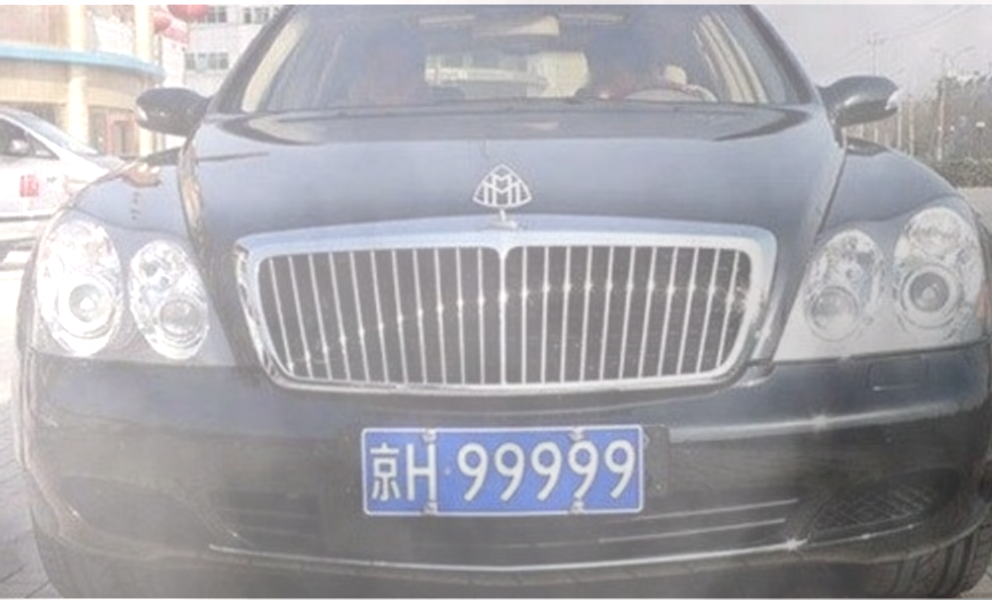}
			
		\end{minipage}
	}%
	\subfigure[$Contrast$]{
		\label{fig:cat:l}
		\begin{minipage}[t]{0.25\linewidth}
			\centering
			\includegraphics[width=1.2in]{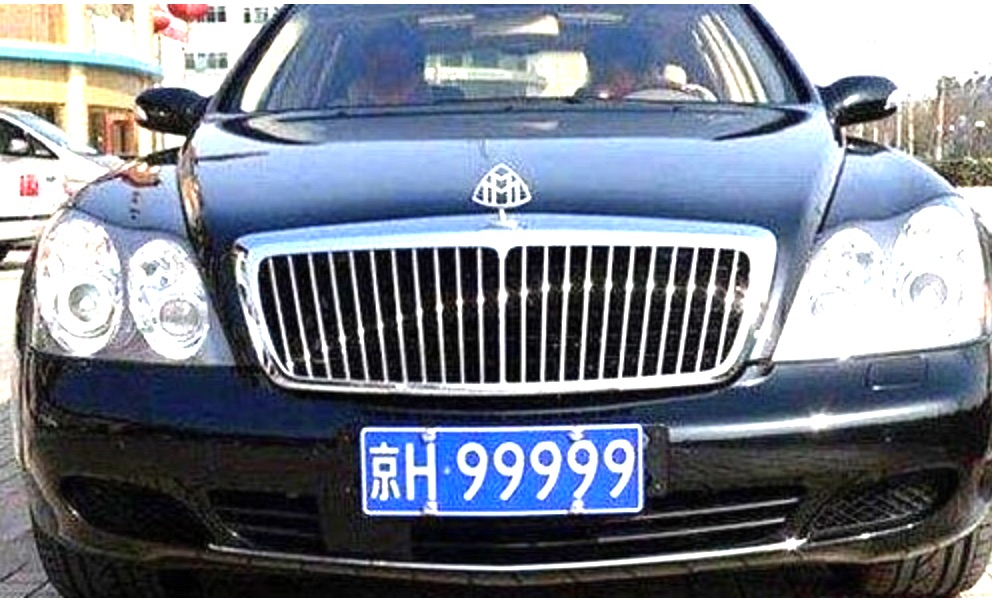}
			
		\end{minipage}
	}%
	
	\subfigure[$Brightness$]{
		\label{fig:cat:m}
		\begin{minipage}[t]{0.25\linewidth}
			\centering
			\includegraphics[width=1.2in]{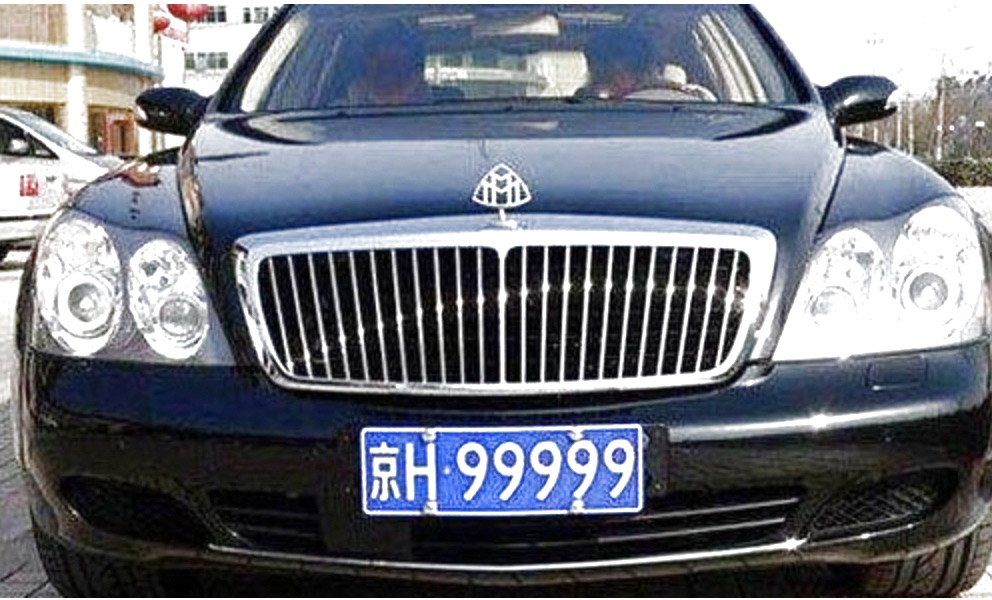}
			
		\end{minipage}
	}%
	\subfigure[$Uniform$]{
		\label{fig:cat:n}
		\begin{minipage}[t]{0.25\linewidth}
			\centering
			\includegraphics[width=1.2in]{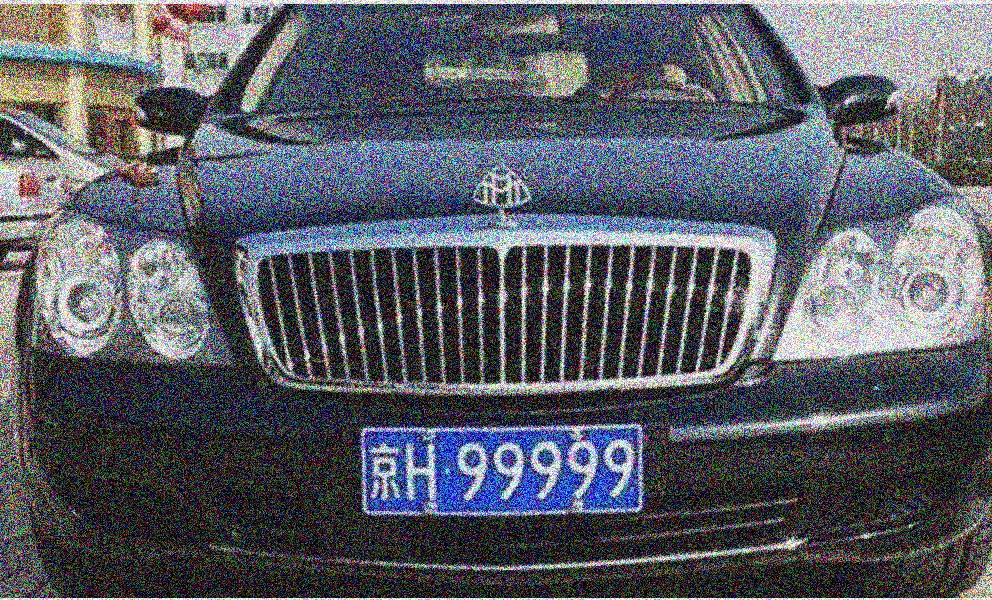}
			
		\end{minipage}
	}%
	\subfigure[$Snowing$]{
		\label{fig:cat:o}
		\begin{minipage}[t]{0.25\linewidth}
			\centering
			\includegraphics[width=1.2in]{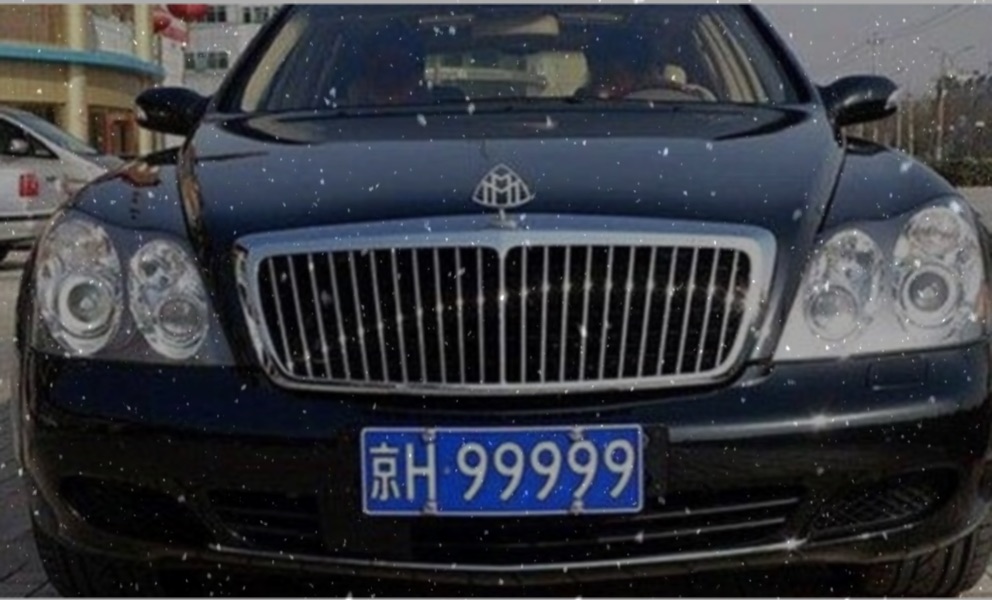}
			
		\end{minipage}
	}%
	\subfigure[$Raining$]{
		\label{fig:cat:p}
		\begin{minipage}[t]{0.25\linewidth}
			\centering
			\includegraphics[width=1.2in]{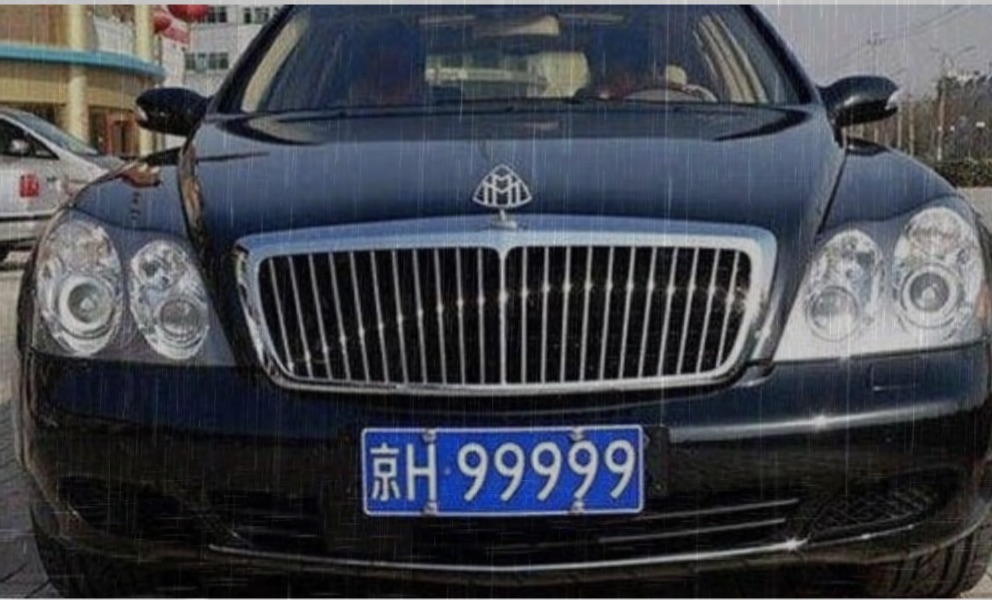}
			
		\end{minipage}
	}%
	
	\centering
	\caption{Illustration of Spatial Attack on a cat image. (a) is origin image, (b) is Gaussian Noise, (c) is Salt-and-Pepper Noise, (d) is Rotation and (e)-(h) is Monochromatization.}
	\label{fig:cat}
\end{figure*}

\end{document}